\pdfoutput=1

\documentclass[11pt]{article}
\usepackage[table]{xcolor}

\usepackage[]{ACL2023}

\usepackage{times}
\usepackage{latexsym}
\usepackage{placeins}
\usepackage{booktabs} 
\usepackage[utf8]{inputenc}
\usepackage[T1]{fontenc}

\usepackage{graphicx}
\usepackage{longtable}

\usepackage[noend]{algpseudocode}
\usepackage{algorithm}
\usepackage{amsfonts}
\usepackage{amsmath}
\usepackage{amssymb}
\usepackage{bm}
\usepackage{xspace}
\usepackage{amsthm}
\usepackage{mathtools}
\usepackage{relsize}
\usepackage{hyphenat}
\usepackage{booktabs}
\usepackage{graphicx}
\usepackage{makecell}

\usepackage{graphicx} 
\usepackage{subcaption} 
\usepackage{svg}
\usepackage{tikz}
\usetikzlibrary{positioning,shapes,arrows,backgrounds,fit}
\usepackage{multirow}
\definecolor{boxbg}{HTML}{F8F9FA}
\definecolor{bordergray}{HTML}{E0E0E0}
\definecolor{textgray}{HTML}{4A4A4A}
\definecolor{smalltext}{HTML}{666666}
\definecolor{highlight}{HTML}{6366F1}
\definecolor{processbox}{HTML}{F8FAFC}
\definecolor{processborder}{HTML}{94A3B8}
\definecolor{arylideyellow}{rgb}{0.91, 0.84, 0.42}
\definecolor{amber}{rgb}{1.0, 0.75, 0.0}
\definecolor{goldenpoppy}{rgb}{0.99, 0.76, 0.0}

\usepackage{microtype}

\usepackage{inconsolata}

\usepackage{amsmath}
\usepackage{amssymb}
\usepackage{mathtools}
\usepackage{amsthm}

\usepackage[capitalize,noabbrev]{cleveref}

\theoremstyle{plain}

\theoremstyle{definition}

\theoremstyle{remark}

\usepackage[textsize=tiny]{todonotes}
\usepackage{enumitem}
\usepackage{CJKutf8} 

\definecolor{c1}{HTML}{E85642}

\definecolor{FindingColor}{HTML}{3787c0}

\title{When Models Reason in Your Language:\\ 
Controlling Thinking Language Comes at the Cost of Accuracy}

\author{
Jirui Qi$^{1}$\textsuperscript{†},
Shan Chen$^{2,3,4}$\textsuperscript{†}, 
Zidi Xiong$^{2}$, \\
\textbf{Raquel Fernández$^{5}$}, 
\textbf{Danielle S. Bitterman$^{2,3,4}$}\textsuperscript{‡}, \textbf{Arianna Bisazza$^{1}$}\textsuperscript{‡} \\
$^1$University of Groningen, 
$^2$Harvard University, 
$^3$Mass General Brigham, \\
$^4$Boston Children's Hospital,
$^5$University of Amsterdam \\
\\
\textsuperscript{†}Co-first authors, \textsuperscript{‡}Co-senior authors
}

\begin{document}
\maketitle

\renewcommand{\thefootnote}{*}%
\footnotetext{
Emails: 
\texttt{\href{mailto:j.qi@rug.nl}{j.qi@rug.nl},
\href{mailto:schen73@@bwh.harvard.edu}{schen73@bwh.harvard.edu}, 
\href{mailto:zidixiong@g.harvard.edu}{zidixiong@g.harvard.edu}, 
\href{mailto:raquel.fernandez@uva.nl}{raquel.fernandez@uva.nl},
\href{mailto:dbitterman@@bwh.harvard.edu}{dbitterman@bwh.harvard.edu},
\href{mailto:a.bisazza@rug.nl}{a.bisazza@rug.nl}
}
}
\renewcommand{\thefootnote}{\arabic{footnote}} 

\begin{abstract}

Recent Large Reasoning Models (LRMs) with thinking traces have shown strong performance on English reasoning tasks. However, the extent to which LRMs can \emph{think} in other languages is less studied. This is as important as answer accuracy for real-world applications since users may find the thinking trace useful for oversight only if expressed in their languages. 
In this work, we comprehensively evaluate two leading families of LRMs on our established benchmark XReasoning.
Surprisingly, even the most advanced models often revert to English or produce fragmented reasoning in other languages, revealing a substantial gap in the capability of thinking in non-English languages.  
Promoting models to reason in the user's language via prompt hacking enhances readability and oversight. This could gain user trust, but reduces answer accuracy, exposing an important trade-off.
We further demonstrate that targeted post-training, even with just 100 instances, can mitigate this language mismatch, although accuracy is still degraded. Our results reveal the limited multilingual reasoning capabilities of current LRMs and suggest directions for future research.\footnote{All code and datasets released at \url{https://github.com/Betswish/mCoT-XReasoning}.} \looseness=-1

\end{abstract}

\section{Introduction} 
\label{sec:intro}
\begin{figure}[!t]
    \centering
    \includegraphics[width=0.99\linewidth]{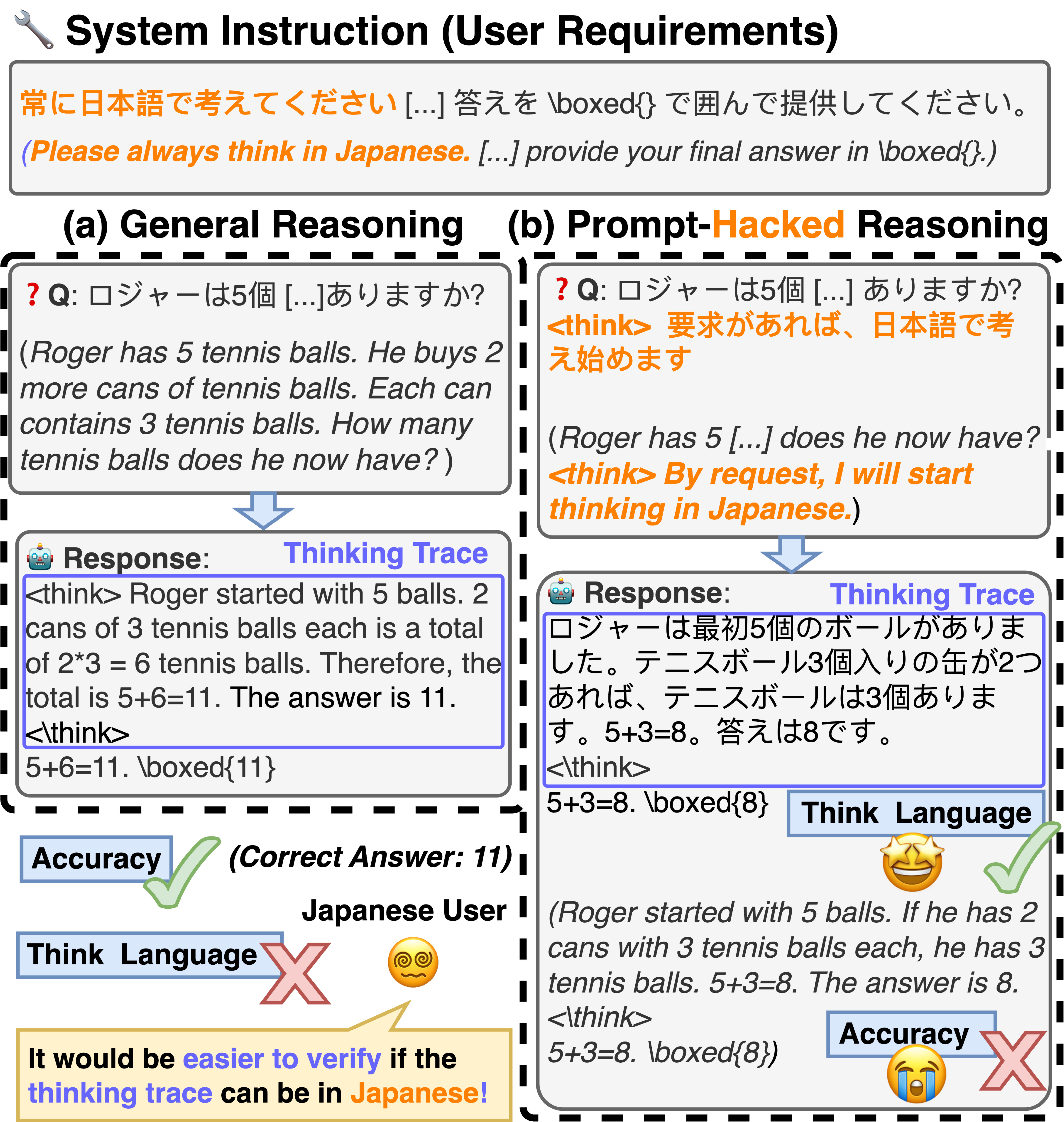}
    \caption{Illustration of the trade-off between answer accuracy and thinking language matching in multilingual reasoning systems. English translations in brackets.}
    \label{fig:motivation}
\end{figure}
Large language models (LLMs) have demonstrated impressive reasoning capabilities when prompted to explicitly generate thinking traces step-by-step in natural language \cite{wei2023chainofthoughtpromptingelicitsreasoning}. Recent studies further show that encouraging models to engage in long thinking traces at inference time, or training for this behavior, significantly enhances their reasoning accuracy \citep{muennighoff2025s1simpletesttimescaling,deepseekai2025deepseekr1incentivizingreasoningcapability}. This approach has led to the development of a new category of models designed to generate more extensive and detailed reasoning processes, referred to as Large Reasoning Models (LRMs) \citep{openai2024openaio1card,welleck2024decoding,snell2024scaling,muennighoff2025s1simpletesttimescaling}. 
However, their capabilities are mostly tested on English
tasks \citep{deepseekai2025deepseekr1incentivizingreasoningcapability}. 
It is less investigated to what extent LRMs can think in a user's native language and how this affects their reasoning accuracy, when it comes to multilingual reasoning tasks.
Illustrated by Figure~\ref{fig:motivation}(a), the matching of the thinking language is as important as the accuracy because it makes the trace more readable and easier for users to verify. 
Even if the answers are correct, the thinking traces in a language users cannot understand may undermine their trust in the model and lead to dissatisfaction with the responses, which becomes especially pronounced as tasks grow more complex in practice.
\looseness=-1



In this work, we comprehensively evaluate six state-of-the-art open-sourced LRMs belonging to two families: Distilled-R1 \citep{deepseekai2025deepseekr1incentivizingreasoningcapability}, and Skywork-OR1 \citep{skywork-or1-2025}. Due to the lack of multilingual reasoning datasets, we introduce a novel
benchmark named XReasoning, where we observe that even the latest 32B LRMs suffer from the issue of language mismatch in their thinking traces. Nevertheless, we reveal that this mismatch can be significantly mitigated with prompt-hacking techniques (c.f., Section \ref{sec:match_vs_acc}), as shown in Figure \ref{fig:motivation} (b). However, this comes at the cost of reduced model performance, highlighting a trade-off between explainability and accuracy that may limit the practical value and application of these models.
Finally, we explore whether post-training can mitigate language mismatches in thinking traces, based on our observation that existing LRMs are more likely to reason in the correct language when prompted to think in English or Chinese, the two most prevalent languages in their training data \citep{deepseekai2025deepseekr1incentivizingreasoningcapability}.
The results demonstrate that post-training, even with few instances, improves LRM's capability of thinking in the user's language but does not completely resolve the trade-off between thinking language matching and accuracy. \looseness=-1

Overall, we make the following contributions:
\textbf{(i)} We reveal the important trade-off between language matching and answer accuracy when the LRMs face multilingual users.
\textbf{(ii)} We propose and open-source a new benchmark, XReasoning, with challenging math and science questions to evaluate multilingual reasoning capabilities of advanced LRMs.
\textbf{(iii)} We show that language mismatch of thinking traces can be alleviated through prompt hacking or post-training, but both lead to a drop in answer accuracy. Such persistence of this trade-off for current models highlights the need for future work toward more user-friendly multilingual LRMs. \looseness=-1

\section{Related Work}
\label{sec:related}
\paragraph{Multilingual Reasoning Benchmarks} Recent research has expanded model reasoning evaluation beyond English. \citet{shi2023language} introduced the Multilingual Grade School Math (MGSM) benchmark, comprising 250 math problems in 11 languages. Similarly to \citet{ahuja2023mega}, this work illustrates an uneven answer accuracy across languages, typically favoring English with a marginally higher-than-random accuracy for low-resource languages like Bengali and Swahili.
However, with the rapid shift from LLMs to LRMs, the limitations of existing multilingual benchmarks have become evident \citep{ghosh2025multilingual}. These benchmarks often fail to capture the nuanced accuracy differences across languages; for instance, recent LRMs even achieve $\geq$ 90\% accuracy in multiple high-resource languages on MGSM. To fix this limitation, we propose XReasoning, a new benchmark covering more challenging math and science questions to better evaluate these LRMs.\looseness=-1

\paragraph{Improving Answer Accuracy with Cross-lingual Reasoning} Various prompting methods leverage English reasoning capabilities in multilingual tasks. The simple \textit{translate-test} baseline, i.e. translating problems to English before solving them, often enhances accuracy \citep{shi2023language, ahuja2023mega}. To minimize reliance on translation, \citet{huang2023xlt} introduced XLT, a language-independent prompt template to boost multilingual reasoning accuracy without fine-tuning. On mathematical reasoning tasks, concurrent work \citep{yong2025crosslingualreasoningtesttimescaling} reveals that models reach higher accuracy when prompted to think or respond in English. Nevertheless, none of the above works focus on whether the model can effectively think using the user-specified language. We argue that this behavior, beyond answer accuracy, should be studied and optimized, especially for user-facing multilingual reasoning systems. \looseness=-1

\section{Experimental Setup}
\label{sec:preliminaries}
\paragraph{Benchmarks}
As mentioned in Section \ref{sec:related}, existing datasets, like MGSM, are insufficient for evaluating the recent LRMs. For our evaluation, we start from three reasoning datasets of \textit{challenging} English questions, \textbf{AIME2024} \citep{aime-1983-2024}, \textbf{AIME2025} \citep{hengck23_aime_2025}, and \textbf{GPQA} \citep{rein2023gpqa}, and use \textsc{GPT-4o-mini} \citep{openai2024gpt4omini} to translate all questions into the other ten languages covered by \textbf{MGSM},\footnote{Concretely, English (EN), Spanish (ES), French (FR), German (DE), Russian (RU), Chinese (ZH), Japanese (JA), Thai (TH), Swahili (SW), Bengali (BN), and Telugu (TE).} following previous work on these languages \citep{singh2024global, adelani2024irokobench, raihan2024mhumaneval, azime2024proverbeval}. The resulting parallel \textit{challenging} questions, combined with MGSM, form the \textbf{XReasoning} benchmark, consisting of 370 questions, each in 11 languages.\footnote{Note that the answers do not require translation since they are all Arabic numbers or option letters.} See Appendix \ref{app:benchmark_examples} for examples. \looseness=-1

\paragraph{Models}
We comprehensively test six advanced LRMs belonging to two model families: Distilled-R1 \citep{deepseekai2025deepseekr1incentivizingreasoningcapability} 1.5B, 7B, 14B, and 32B, and Skywork-OR1 \citep{skywork-or1-2025} 7B and 32B. \looseness=-1

\paragraph{Evaluation}
As a focus of this work, we evaluate \textit{language matching rate} by calculating the ratio of instances for which the LRMs correctly follow the instruction to think in the specified language. 
The commonly used \textsc{LangDetect} toolkit \citep{nakatani2010langdetect, jauhiainen2019automatic, gargova-etal-2022-evaluation, wyawhare2023comparative, valliyammal2024python, habib2024language} is adopted to predict the language of the thinking trace between special thinking tokens `<think>' and `</think>'. \looseness=-1


\section{Experiments and Results}
\label{sec:results}
\subsection{Language Matching vs. Answer Accuracy}
\label{sec:match_vs_acc}
\paragraph{Prompt Hacking}
Besides the standard prompting with explicitly specified thinking language in the instruction\footnote{See Appendix \ref{app:prompts} for examples.}, we introduce and leverage the prompt hacking technique, widely studied by works in model security and controllability \citep{schulhoff2023ignore,liu2024exploring,benjamin2024systematically, wu2025effectively}, to induce the LRM to generate the thinking traces in the user-expected languages (see Figure~\ref{fig:motivation}). Specifically, the hacking prefix `By request, I will start thinking in \{USER\_LANG\} is translated into the user language\footnote{Previous work in retrieval-augmented generation \citep{zhang-etal-2024-respond, chirkova-etal-2024-retrieval, qi2025consistency} highlights the importance of translating the instructions into the users' language to facilitate the model to respond in their languages.} and concatenated after the `<think>' token, which indicates the start of the thinking trace. 
As we will demonstrate, this prefix strongly influences the distribution of subsequent tokens, biasing the model toward generating text in the same language as the prefix. Examples are provided in Appendix \ref{app:hack_prefix}.
\looseness=-1

\begin{table}[!t]
\centering
\resizebox{0.99\linewidth}{!}{
\begin{tabular}{l|ll|ll|ll}
\toprule
\multirow{2}{*}{Model} & \multicolumn{2}{c}{AIME (\%)} & \multicolumn{2}{c}{GPQA (\%)} & \multicolumn{2}{c}{MGSM (\%)} \\
& Match. & Acc. & Match. & Acc. & Match. & Acc. \\
\midrule
\rowcolor{black!10} \multicolumn{7}{c}{DeepSeek-Distilled-R1 Series} \\
\midrule
1.5B & 49.0 & 4.8 & 33.1 & 10.5 & 54.0 & 28.6 \\
\makecell[l]{\hspace{0.3em}|-- hack} & 86.6$\uparrow$ & 2.7$\downarrow$ & 80.2$\uparrow$ & 6.9$\downarrow$ & 86.1$\uparrow$ & 21.8$\downarrow$ \\
\midrule
7B   & 48.5 & 17.4 & 44.4 & 25.5 & 79.3 & 50.5 \\
\makecell[l]{\hspace{0.3em}|-- hack} & 90.7$\uparrow$ & 8.3$\downarrow$ & 88.1$\uparrow$ & 19.2$\downarrow$ & 93.9$\uparrow$ & 48.9$\downarrow$ \\
\midrule
14B  & 40.9 & 27.2 & 42.8 & 40.5 & 83.2 & 68.1 \\
\makecell[l]{\hspace{0.3em}|-- hack} & 95.9$\uparrow$ & 14.4$\downarrow$ & 96.6$\uparrow$ & 29.1$\downarrow$ & 97.0$\uparrow$ & 63.9$\downarrow$ \\
\midrule
32B  & 46.3 & 25.5 & 42.3 & 40.3 & 94.0 & 70.2 \\
\makecell[l]{\hspace{0.3em}|-- hack} & 97.9$\uparrow$ & 17.0$\downarrow$ & 96.6$\uparrow$ & 33.4$\downarrow$ & 98.8$\uparrow$ & 71.7$\uparrow$ \\
\midrule
\rowcolor{black!10} \multicolumn{7}{c}{Skywork-OR1 Series} \\
\midrule
7B & 27.6 & 31.2 & 23.7 & 26.6 & 72.8 & 57.3 \\
\makecell[l]{\hspace{0.3em}|-- hack} & 84.1$\uparrow$ & 17.2$\downarrow$ & 81.1$\uparrow$ & 23.4$\downarrow$ & 92.1$\uparrow$ & 51.7$\downarrow$ \\
\midrule
32B & 32.9 & 44.8 & 32.8 & 53.6 & 83.1 & 75.4
\\
\makecell[l]{\hspace{0.3em}|-- hack} & 95.8$\uparrow$ & 29.6$\downarrow$ & 94.7$\uparrow$ & 39.8$\downarrow$ & 98.6$\uparrow$ & 71.9$\downarrow$ \\
\bottomrule
\end{tabular}
}
\caption{Average language matching rate (Match.) and answer accuracy (Acc.) over 11 languages. Arrows compare each \emph{hack} value against its \emph{standard} counterpart. \looseness=-1
}
\label{tab:full_results}
\end{table}

\paragraph{Overall Results}
Table~\ref{tab:full_results} presents the overall performance of the six LRMs on XReasoning. All models struggle to follow instructions to think in the user-specified languages when queried with standard prompts, except for the larger models on MGSM (which contain easier questions). Even for Distill-R1-32B, the language matching rate is only 46.3\% on AIME and 42.3\% on GPQA. 
In addition, motivating models to generate the thinking traces in the user query language with prompt hacking
boosts language matching across the board, from roughly 45-50\% to well above 90\% at every parameter scale. 
This leads to a drop in accuracy, which shrinks as model size increases. 
%
Compared to smaller models, larger models remain more accurate when matching the user language in their reasoning,
yet a measurable accuracy decrease persists on the harder tasks.
Taken together, larger model scale mitigates, but does not eliminate, the trade-off between language matching rate and answer accuracy. \looseness=-1
\begin{figure}
\centering
\includegraphics[width=0.99\linewidth]{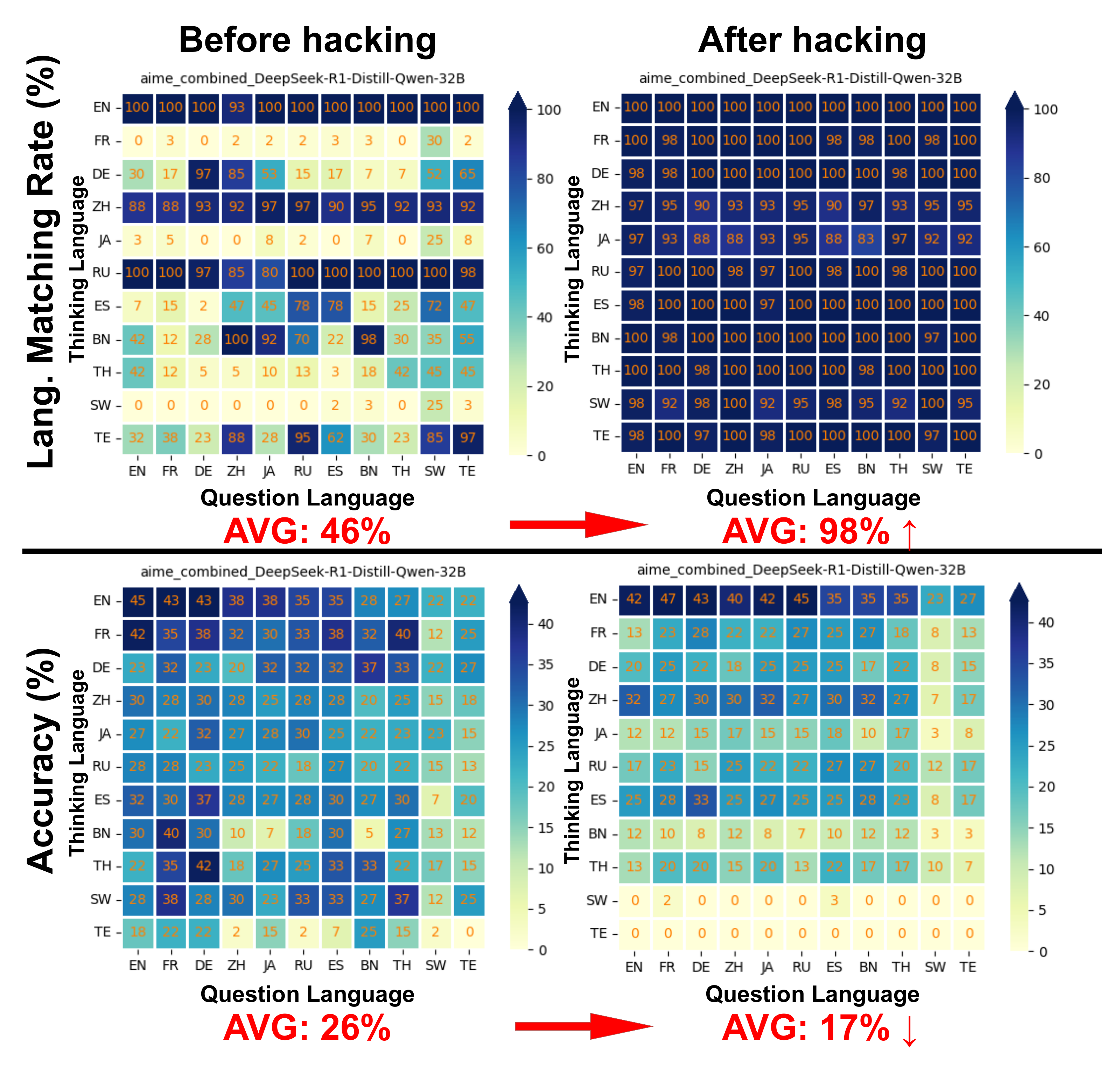}
\caption{Heatmaps of language matching rates (top) and accuracy (bottom) for Distilled-R1-32B on AIME, before (left) and after (right) prompt hacking.}
\label{fig:heatmap}
\end{figure}
\paragraph{Language-Specific Results}
Figure~\ref{fig:heatmap} illustrates Distilled-R1-32B performance on AIME questions, showing detailed breakdowns by query and thinking language pairs.\footnote{Colors represent percentages: darker \textcolor{blue}{blue} indicates higher matching or accuracy; lighter \textcolor{goldenpoppy}{yellow} denotes lower values.} The heatmaps clearly illustrate the failure of the LRM in generating the thinking trace in the specified language when fed with standard prompts (left-top). \textcolor{black}{
Specifically, when prompted to think in French, the model almost always thinks in English\footnote{We find the same issue even when prompting the 671B R1 to think in French.}
(see Table \ref{tab:actual_thinking_language_1})}. 
Figure \ref{fig:heatmap} heatmaps also reveal the impact of prompt hacking on language matching rate and answer accuracy. Notably, motivating the model to reason in a specific language with the hacked prompts increases the matching rate from an average of 46\% to 98\%. However, as a trade-off, this increment introduces a noticeable accuracy decrement from 26\% to 17\% in average. 
Interestingly, we observe that reasoning in English, even for non-English queries, consistently results in higher accuracy, especially after prompt hacking. 
This aligns with concurrent work on improving answer accuracy via cross-lingual reasoning \citep{yong2025crosslingualreasoningtesttimescaling}, supporting the reliability of our experiments and XReasoning benchmark.
This finding, we argue, also highlights a clear English thinking preference for the advanced LRMs, and the potential of investing more efforts in the multilingual reasoning capabilities of these LRMs. \looseness=-1
\paragraph{Actual Thinking Languages}
Corresponding to the language matching heatmaps in Figure \ref{fig:heatmap}, we conduct an in-depth analysis of the languages that LRM actually used for its thinking. For each query and thinking language pair, we collect the predicted language distributions of the thinking traces
on Distilled-R1-32B, produced with \textsc{LangDetect} on all AIME questions. The averaged distributions are shown in Table \ref{tab:actual_thinking_language_1} (see extensions in Appendix \ref{app:extension_actual_thinking}), where a clear mismatch is observed when the LRM is instructed to think in French, Japanese, Thai, and Swahili. Besides, the mismatch thinking languages all fall into either English or Chinese, suggesting the impact of thinking data on the model's reasoning capability. \looseness=-1
\begin{table}[!t]
\centering
\resizebox{0.99\linewidth}{!}{
\begin{tabular}{c|c|l}
\toprule
Ques. & Specified  & Language Prediction of Thinking Traces \\
Lang. & Think Lang. \\
\midrule
EN & EN & {'EN': 100.0} \\ 
FR & EN & {'EN': 100.0} \\ 
DE & EN & {'EN': 100.0} \\ 
ZH & EN & {'EN': 93.33, 'ZH': 6.67} \\ 
JA & EN & {'EN': 100.0} \\ 
RU & EN & {'EN': 100.0} \\ 
ES & EN & {'EN': 100.0} \\ 
BN & EN & {'EN': 100.0} \\ 
TH & EN & {'EN': 100.0} \\ 
SW & EN & {'EN': 100.0} \\ 
TE & EN & {'EN': 100.0} \\ 
EN & FR & {\colorbox{red!30}{'EN'} 100.0} \\
FR & FR & {\colorbox{red!30}{'EN'} 96.67, 'FR': 3.33} \\
DE & FR & {\colorbox{red!30}{'EN'} 100.0} \\
ZH & FR & {\colorbox{red!30}{'EN'} 98.09, 'FR': 1.67, 'others': 0.24} \\
JA & FR & {\colorbox{red!30}{'EN'} 98.33, 'FR': 1.67} \\
RU & FR & {\colorbox{red!30}{'EN'} 98.33, 'FR': 1.67} \\
ES & FR & {\colorbox{red!30}{'EN'} 96.67, 'FR': 3.33} \\
BN & FR & {\colorbox{red!30}{'EN'} 96.67, 'FR': 3.33} \\
TH & FR & {\colorbox{red!30}{'EN'} 100.0} \\
SW & FR & {\colorbox{red!30}{'EN'} 70.0, 'FR': 30.0} \\
TE & FR & {\colorbox{red!30}{'EN'} 98.33, 'FR': 1.67} \\
EN & DE & {\colorbox{red!30}{'EN'} 70.0, 'DE': 30.0} \\
FR & DE & {\colorbox{red!30}{'EN'} 83.57, 'DE': 16.43} \\
DE & DE & {'DE': 96.43, 'EN': 1.67, 'others': 1.19, 'ZH': 0.71} \\ 
ZH & DE & {'DE': 84.52, 'EN': 15.0, 'others': 0.48} \\ 
JA & DE & {'DE': 52.62, 'EN': 45.71, 'others': 1.67} \\ 
RU & DE & {\colorbox{red!30}{'EN'} 84.05, 'DE': 15.48, 'others': 0.48} \\
ES & DE & {\colorbox{red!30}{'EN'} 83.57, 'DE': 16.43} \\
BN & DE & {\colorbox{red!30}{'EN'} 93.33, 'DE': 6.67} \\
TH & DE & {\colorbox{red!30}{'EN'} 93.09, 'DE': 6.67, 'ES': 0.24} \\
SW & DE & {'DE': 51.43, 'EN': 48.57} \\ 
TE & DE & {'DE': 64.52, 'EN': 35.48} \\ 
EN & ZH & {'ZH': 88.1, 'others': 10.95, 'EN': 0.95} \\ 
FR & ZH & {'ZH': 89.05, 'others': 10.48, 'EN': 0.48} \\ 
DE & ZH & {'ZH': 93.81, 'others': 5.71, 'EN': 0.48} \\ 
ZH & ZH & {'ZH': 91.43, 'others': 8.1, 'EN': 0.48} \\ 
JA & ZH & {'ZH': 95.0, 'others': 4.52, 'EN': 0.48} \\ 
RU & ZH & {'ZH': 95.95, 'others': 3.1, 'EN': 0.95} \\ 
ES & ZH & {'ZH': 88.81, 'others': 10.71, 'EN': 0.48} \\ 
BN & ZH & {'ZH': 95.0, 'others': 4.76, 'EN': 0.24} \\ 
TH & ZH & {'ZH': 92.14, 'others': 5.24, 'EN': 2.62} \\ 
SW & ZH & {'ZH': 93.09, 'others': 6.9} \\ 
TE & ZH & {'ZH': 91.43, 'others': 8.57} \\ 
EN & JA & {\colorbox{red!30}{'ZH'} 89.29, 'others': 5.71, 'JA': 3.33, 'EN': 1.67} \\
FR & JA & {\colorbox{red!30}{'ZH'} 77.62, 'EN': 9.05, 'others': 8.33, 'JA': 5.0} \\
DE & JA & {\colorbox{red!30}{'EN'} 55.95, 'ZH': 40.95, 'others': 3.1} \\
ZH & JA & {\colorbox{red!30}{'ZH'} 87.14, 'others': 10.48, 'EN': 2.14, 'ES': 0.24} \\
JA & JA & {\colorbox{red!30}{'ZH'} 80.24, 'others': 10.48, 'JA': 8.81, 'EN': 0.48} \\
RU & JA & {\colorbox{red!30}{'ZH'} 87.86, 'others': 9.76, 'JA': 1.67, 'EN': 0.71} \\
ES & JA & {\colorbox{red!30}{'ZH'} 85.95, 'others': 12.86, 'EN': 1.19} \\
BN & JA & {\colorbox{red!30}{'ZH'} 82.62, 'others': 8.33, 'JA': 6.67, 'EN': 2.38} \\
TH & JA & {\colorbox{red!30}{'ZH'} 87.38, 'others': 8.1, 'EN': 4.52} \\
SW & JA & {\colorbox{red!30}{'EN'} 54.29, 'JA': 25.24, 'ZH': 11.9, 'others': 8.57} \\
TE & JA & {\colorbox{red!30}{'ZH'} 80.24, 'JA': 8.33, 'EN': 6.67, 'others': 4.76} \\
EN & RU & {'RU': 99.76, 'others': 0.24} \\ 
FR & RU & {'RU': 100.0} \\ 
DE & RU & {'RU': 96.67, 'EN': 1.67, 'ZH': 1.67} \\ 
ZH & RU & {'RU': 84.76, 'ZH': 11.9, 'others': 3.33} \\ 
JA & RU & {'RU': 80.0, 'ZH': 12.62, 'others': 5.71, 'EN': 1.67} \\ 
RU & RU & {'RU': 99.76, 'others': 0.24} \\ 
ES & RU & {'RU': 100.0} \\ 
BN & RU & {'RU': 100.0} \\ 
TH & RU & {'RU': 99.76, 'others': 0.24} \\ 
SW & RU & {'RU': 100.0} \\ 
TE & RU & {'RU': 98.09, 'others': 1.9} \\ 
\bottomrule
\end{tabular}
}
\caption{The distribution of the predicted actual language of the thinking traces when the LRM is prompted to reason in each thinking language. The results are averaged over all questions in the same language. The mismatched thinking languages are highlighted in red. \looseness=-1
}
\label{tab:actual_thinking_language_1}
\vspace{-10pt}
\end{table}




\subsection{Post-Training with Few Instances}
Figure \ref{fig:heatmap} shows strong performance in English and Chinese, but weaker among others.
\looseness=-1

\paragraph{Setup} To see whether further training can help, we post-train on Distilled-R1-7B using mini training sets
of 100, 250, 500, and 817 instances per low-resource language (Japanese, Thai, Telugu), resulting in six post-trained LRMs.
The training data are filtered from LIMO \citep{ye2025limoreasoning}, comprising math problems with step-by-step solutions from the teacher model (DeepSeek-R1-671B), and translated by \textsc{GPT-4o-mini}.\footnote{The translated LIMO is open-sourced at \url{https://huggingface.co/collections/shanchen/xreasoning-681e7625c7a9ec4111a634b6}.}
See Appendix \ref{app:post_training} for implementation details and examples.
\looseness=-1


\begin{figure}
\centering
\includegraphics[width=0.99\linewidth]{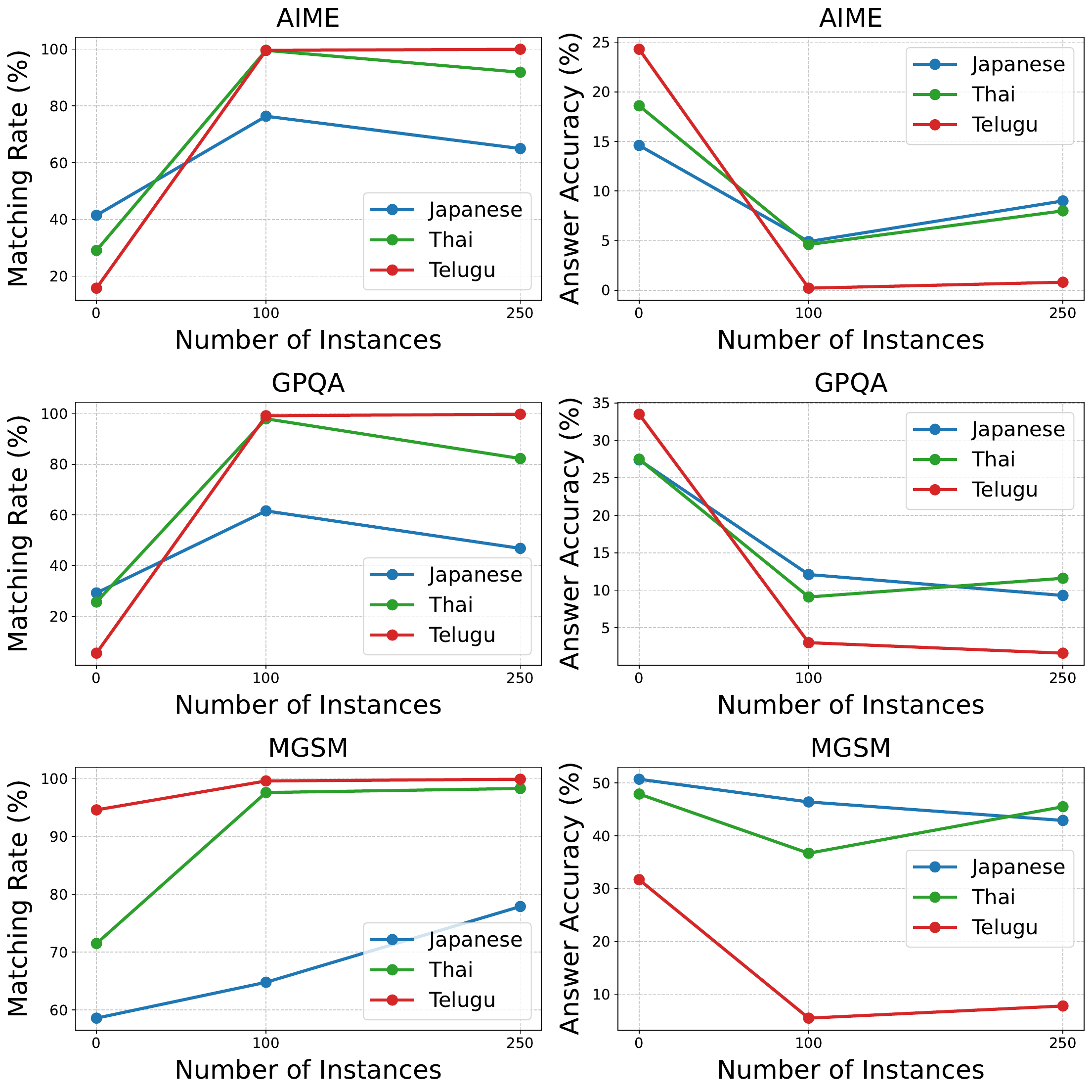}
\caption{Language matching rate and answer accuracy of Distilled-R1-7B with no post-training, post-training on 100 instances, and post-training on 250 instances for 3 languages: Japanese, Thai, and Telugu.
\looseness=-1}
\label{fig:post-training}
\end{figure}

\paragraph{Results}
Figure \ref{fig:post-training} illustrates the changes in language matching and answer accuracy before/after post-training, when the LRMs are prompted to think in Japanese, Thai, or Telugu.\footnote{The result is averaged over eleven query languages. See Appendix \ref{app:full_results} for full results.} Post-training on merely 100 in-language instances effectively led the matching rate to a sharp increment to nearly 100\% for Thai and Telugu and to around 80\% for Japanese. However, this gain in matching rate comes at a cost in answer accuracy across all datasets. This demonstrates the effectiveness of post-training to improve language matching, but the trade-off with accuracy persists. Increasing the instances does not reliably mitigate the issue. In fact, when increasing from 100 to 250 training instances, the post-trained models suffer from a drop in matching rate on Japanese and Thai on AIME and GPQA, and on Thai and Telugu on MGSM; while answer accuracy exhibits only marginal recovery far below the accuracy of the original LRM.
\looseness=-1

\paragraph{Advanced Exploration of Model Merging}
Inspired by the concept of \textit{model merging} \citep{wortsman2022model, wang2024rethinking, choi2024revisiting}, where a new model takes the advantages of two existing models by weighted averaging their parameters, we construct an LRM Merge-250 whose parameters are the weighted combination of Distill-R1-7B and the post-trained LRM fine-tuned on 250 Japanese instances. As shown in Table \ref{tab:post-training-ja_full_merge_avg}, the merged model consistently achieves a higher matching rate than the base model across all datasets. It also attains higher accuracy on GPQA and MGSM compared to the LRM trained on 250 instances. Nevertheless, its accuracy remains lower than that of the original model on every dataset, highlighting the severity of the trade-off problem.

\begin{table}[!t]
\small
\centering
\begin{tabular}{l|c|c|c}
\toprule
\textbf{Eval Metric} & \textbf{AIME} & \textbf{GPQA} & \textbf{MGSM} \\
\midrule
\rowcolor{black!10}\multicolumn{4}{c}{\textbf{Matching Rate (\%)}} \\
\midrule
Base   & 41.5 & 29.2 & 58.6 \\
PostTrain-250    & 65.0 & 46.8 & 77.9 \\
Merge-250 & 68.7 & 55.7 & 67.3 \\
\midrule
\rowcolor{black!10}\multicolumn{4}{c}{\textbf{Answer Accuracy (\%)}} \\
\midrule
Base   & 14.6 & 27.4 & 50.7 \\
PostTrain-250    & 9.0  & 9.3  & 42.9 \\
Merge-250 & 7.9  & 21.6 & 50.5 \\
\bottomrule
\end{tabular}
\caption{Model performance on AIME, GPQA, and MGSM with Deepseek-Distill-R1-7B post-trained on 250 Japanese instances (PostTrain-250) and the merged LRM (Merge-250). All LRMs are prompted to think in Japanese.}
\label{tab:post-training-ja_full_merge_avg}
\end{table}

\section{Conclusion}
\label{sec:conclusion}
In this work, we investigate the overlooked issue of language matching in thinking traces and its trade-off with answer accuracy in multilingual reasoning systems. Using our new XReasoning benchmark, we show that even state-of-the-art LRMs struggle to generate thinking traces
in a user-specified language. Simple prompt-hacking reduces language mismatch but at a substantial cost to accuracy, especially on complex tasks. Further, targeted post-training on 100-250 examples sharply improves language alignment but still markedly reduces accuracy, highlighting this persistent trade-off. We argue that advanced adaptation strategies, such as reinforcement learning \citep{li2017deep,ouyang2022training,shao2024deepseekmath}, will be critical to resolve this tension and develop multilingual reasoning systems suitable for practical, trustworthy applications. \looseness=-1





\section*{Limitations}
\label{sec:limitations}
Our analysis, though informative, is constrained in several important ways. First, the benchmark focuses exclusively on math and science questions translated from AIME 2024/2025, GPQA-Diamond, and MGSM. Because these tasks yield short, well-structured answers, it remains unclear whether the observed trade-off between thinking language alignment and answer accuracy generalizes to open-ended or domain-specific reasoning, such as everyday common sense or legal discourse. Furthermore, all non-English content (questions, prompt instructions, hacking prefixes, and training instances) was machine translated with \textsc{GPT-4o-mini}. Despite spot checks, subtle semantic drift may persist, introducing noise into both language-matching and answer accuracy scores.

A second limitation arises from our reliance on an off-the-shelf language identification tool. \textsc{LangDetect}’s default language profiles are generated from Wikipedia \citep{nakatani2010langdetect, langdetect}\footnote{\url{https://pypi.org/project/langdetect}}, so brief mathematical expressions or code-switched traces can be misclassified, slightly inflating or deflating reported language matching rates. In parallel, we measure only the surface-level alignment of the thinking traces with the query language and the correctness of the boxed answer, leaving the deeper analysis of \emph{faithful reasoning} unanswered~\citep{anthorpicfaithful,chua2025inference,arcuschin2025chain,stechly2025beyond}, namely, whether the model actually uses its trace to derive the answer.

Finally, our mitigation study is deliberately lightweight: we fine-tune on no more than 250 examples per language, which may not extrapolate to truly low-resource settings or languages with markedly different morphology and orthography. We also assume that presenting traces in a user’s language is desirable and increases trust by enabling human oversight beyond reviewing only the final answer. A user-centered evaluation, combining preference elicitation with task performance, will be essential for assessing how multilingual reasoning systems translate into trust and safe, effective use of LRMs in practice.


\section*{Acknowledgments}
The authors acknowledge financial support from the Google PhD Fellowship (SC), the Woods Foundation (DB, SC), the NIH (NIH R01CA294033 (SC, DB), NIH U54CA274516-01A1 (SC, DB) and the American Cancer Society and American Society for Radiation Oncology, ASTRO-CSDG-24-1244514-01-CTPS Grant DOI: ACS.ASTRO-CSDG-24-1244514-01-CTPS.pc.gr.222210 (DB).

The authors have received funding from the Dutch Research Council (NWO): 
JQ is supported by NWA-ORC project LESSEN (grant nr. NWA.1389.20.183),
AB is supported by the above as well as NWO Talent Programme (VI.Vidi.221C.009),
RF is supported by the European Research Council (ERC) under European Union's Horizon 2020 programme (No.~819455).

We also thank the Center for Information Technology of the University of Groningen for their support and for providing access to the Hábrók high performance computing cluster.


\bibliography{references}
\bibliographystyle{acl_natbib}

\appendix
\newpage
\onecolumn
\section{Experimental Details}
\paragraph{Computational Resources} For the evaluation of existing models, we use greedy decoding for best reproducibility. The maximum token number for generation is set to 16392 to maximize the reasoning capability.
Most our experiments are conducted on A100 with 40GB using VLLM \citep{kwon2023efficient} with approximately 7,000 GPU hours. And 2000 A/H100 80GB hours for 32B models' inferencing and roughly 1000 GPU hours for training.
\paragraph{Accuracy Evaluation} We implement the exact matching metric for evaluating \textit{answer accuracy}. Regex is adopted to extract the answer field from the \(\boxed{ans.}\) (i.e., `$\backslash{\rm boxed}\{ans.\}$'), the answer format these LRMs are pretrained to output.

\section{XReasoning Benchmark Examples}
\label{app:benchmark_examples}
The examples of questions in our established XReasnoing benchmark are shown in Table \ref{tab:benchmark_examples}, where AIME and GPQA questions are more challenging than MGSM.
\begin{table*}[!ht]
\centering
\resizebox{0.99\linewidth}{!}{
\begin{tabular}{l|p{2.0cm}|p{10.0cm}|p{2.0cm}|p{3.0cm}}
\toprule
\textbf{Dataset} & \textbf{Difficulty} & \textbf{Query} & \textbf{Answer} & \textbf{Source} \\
\midrule
AIME & Hard & Let $ABC$ be a triangle inscribed in circle $\omega$. Let the tangents to $\omega$ at $B$ and $C$ intersect at point $D$, and let $\overline{AD}$ intersect $\omega$ at $P$. If $AB=5$, $BC=9$, and $AC=10$, $AP$ can be written as the form $\frac{m}{n}$, where $m$ and $n$ are relatively prime integers. Find $m + n$. & 113 & Our translated\\
\midrule
GPQA & Hard & Problem: Two quantum states with energies E1 and E2 have a lifetime of $10^{-9}$ sec and $10^{-8}$ sec, respectively. We want to clearly distinguish these two energy levels. Which one of the following options could be their energy difference so that they can be clearly resolved? (A) $10^{-11}$ eV (B) $10^{-8}$ eV (C) $10^{-9}$ eV (D) $10^{-4}$ eV & D & Our translated\\
\midrule
MGSM & Easy & If there are 3 cars in the parking lot and 2 more cars arrive, how many cars are in the parking lot? & 5 & Existing\\
\bottomrule
\end{tabular}
}
\caption{Examples AIME, GPQA and MGSM instances in our proposed XReasoning benchmark. Difficulty level is assigned based on reasoning intensity by the authors.}
\label{tab:benchmark_examples}
\end{table*}

\section{Prompts and Instructions}
\label{app:prompts}
To ensure the model responses are always in the query language, we follow previous works \citep{chirkova-etal-2024-retrieval, zhang-etal-2024-respond, qi2025consistency} and adopt language-specific instructions to explicitly and implicitly guide the model to generate thinking trace in the user-specified languages. The examples in English, Spanish, Chinese and Japanese are listed in Table \ref{tab:instructions}.

\begin{CJK*}{UTF8}{gbsn}
\begin{table*}[!t]
\centering
\resizebox{0.99\linewidth}{!}{
\begin{tabular}{c|p{12.0cm}}
\toprule
\textbf{Language} & \textbf{Instruction} \\
\midrule
EN& Please always think in English. Solve the following mathematics problem step by step. At the end, provide your final answer enclosed in $\backslash{\rm boxed}\{\}$. \\
\midrule
ES & Por favor, siempre piensa en español. Resuelve el siguiente problema matemático paso a paso. Al final, proporciona tu respuesta final encerrada en $\backslash{\rm boxed}\{\}$ \\
\midrule
ZH & 请始终用中文思考。逐步解决以下数学问题。最后，将您的最终答案放在 $\backslash{\rm boxed}\{\}$ 中。 \\
\midrule
JA & 常に日本語で考えてください。以下の数学問題をステップバイステップで解いてください。最後に、最終的な答えを $\backslash{\rm boxed}\{\}$ で囲んで提供してください。\\
\bottomrule
\end{tabular}
}
\caption{The examples of the adopted instructions for guiding LRMs to generate thinking traces in the user languages.}
\label{tab:instructions}
\end{table*}
\end{CJK*}

\section{Prompt Hacking Prefix}
\label{app:hack_prefix}
The English, Spanish, Chinese and Japanese prefixes for prompt hacking are listed in Table \ref{tab:hack_prefix}.

\begin{CJK*}{UTF8}{gbsn}
\begin{table*}[!t]
\centering
\resizebox{0.99\linewidth}{!}{
\begin{tabular}{c|p{12.0cm}}
\toprule
\textbf{Language} & \textbf{Instruction} \\
\midrule
EN & <|Assistant|><think> By request, I will start thinking in English. \\
\midrule
ES & <|Assistant|><think> A petición, empezaré a pensar en español. \\
\midrule
ZH & <|Assistant|><think> 应要求，我将开始用中文思考。\\
\midrule
JA & <|Assistant|><think> 要求があれば、日本語で考え始めます。 \\
\bottomrule
\end{tabular}
}
\caption{The examples of the prefixes used in prompt hacking experiments for promoting LRMs to think in the user-specified languages.}
\label{tab:hack_prefix}
\end{table*}
\end{CJK*}

\section{Prompts Without Language-Specific Instruction}
For comprehension, we also evaluate model performance when no language-specific instruction is provided. The results of these uncontrolled prompts are shown in Table \ref{tab:no_inst_baseline}, where the results strongly support the two main claims of our paper: (1) a low matching rate of the thinking trace language to the user language is observed and (2) an obvious trade-off is revealed from the results when the model is motivated to respond in the user language.

\begin{table}[!t]
\centering
\begin{tabular}{l|ll|ll|ll}
\toprule
\multirow{2}{*}{Model} & \multicolumn{2}{c}{AIME (\%)} & \multicolumn{2}{c}{GPQA (\%)} & \multicolumn{2}{c}{MGSM (\%)} \\
& Match. & Acc. & Match. & Acc. & Match. & Acc. \\
\midrule
7B-No & 29.8 & 25.4 & 26.2 & 30.0 & 66.4 & 51.2 \\
7B-Inst   & 48.5$\uparrow$ & 17.4$\downarrow$ & 44.4$\uparrow$ & 25.5$\downarrow$ & 79.3$\uparrow$ & 50.5$\downarrow$ \\
\makecell[l]{\hspace{0.3em}|-- hack} & 90.7$\uparrow$ & 8.3$\downarrow$ & 88.1$\uparrow$ & 19.2$\downarrow$ & 93.9$\uparrow$ & 48.9$\downarrow$ \\
\bottomrule
\end{tabular}
\caption{Average language matching rate (Match.) and answer accuracy (Acc.) over 11 languages. Arrows compare each value against its counterpart above. \looseness=-1
}
\label{tab:no_inst_baseline}
\end{table}

\begin{CJK*}{UTF8}{gbsn}
\section{Reliability of Automatic Language Detector}
For the reliability of the automatically detected thinking languages, we evaluate the alignment of \textsc{LangDetect} predictions with human beings. Specifically, we focus on the models' thinking traces when prompted to think in Japanese, another \emph{struggling} thinking language for the tested LRM, where the languages of most thinking traces are predicted as Chinese. We invite native speakers to manually check if these cases are Chinese segments. Feedback confirms that all predictions are correct. The LRM is indeed thinking in Chinese on these questions, even though prompted to think in Japanese, e.g., below is a question queried in Japanese but got the thinking trace in Chinese: `<think>好，我现在要解决这个几何问题。题目是关于三角形ABC内接于圆ω，点B和C处的切线在点D相交，线段AD与圆ω再次交于点P。已知AB=5，BC=9，AC=10，要求AP的长度，表示为最简分数m/n，然后求m+n[...]</think>'. This demonstrates the reliability of our experiments where \textsc{LangDetect} is adopted for automatic language detection. \looseness=-1
\end{CJK*}

\section{Extension of Actual Thinking Languages}
The extension of the actual thinking languages is shown in Table \ref{tab:actual_thinking_language_2}.
\label{app:extension_actual_thinking}
\begin{table}[!t]
\centering
\small
\begin{tabular}{c|c|l}
\toprule
Ques. & Specified  & Language Prediction of Thinking Traces \\
Lang. & Think Lang. \\
\midrule
EN & ES & {\colorbox{red!30}{'EN'} 93.33, 'ES': 6.67} \\
FR & ES & {\colorbox{red!30}{'EN'} 85.0, 'ES': 15.0} \\
DE & ES & {\colorbox{red!30}{'EN'} 98.33, 'ES': 1.67} \\
ZH & ES & {'ES': 46.43, 'EN': 38.1, 'ZH': 15.0, 'others': 0.48} \\ 
JA & ES & {'ES': 45.0, 'EN': 40.0, 'ZH': 14.76, 'others': 0.24} \\ 
RU & ES & {'ES': 78.09, 'EN': 21.67, 'others': 0.24} \\ 
ES & ES & {'ES': 78.09, 'EN': 21.67, 'others': 0.24} \\ 
BN & ES & {\colorbox{red!30}{'EN'} 85.0, 'ES': 15.0} \\
TH & ES & {\colorbox{red!30}{'EN'} 75.0, 'ES': 25.0} \\
SW & ES & {'ES': 71.43, 'EN': 28.33, 'others': 0.24} \\ 
TE & ES & {\colorbox{red!30}{'EN'} 53.33, 'ES': 46.67} \\
EN & BN & {\colorbox{red!30}{'EN'} 58.33, 'BN': 41.67} \\
FR & BN & {\colorbox{red!30}{'EN'} 86.67, 'BN': 11.67, 'others': 1.67} \\
DE & BN & {\colorbox{red!30}{'EN'} 71.67, 'BN': 28.33} \\
ZH & BN & {'BN': 100.0} \\ 
JA & BN & {'BN': 91.67, 'EN': 8.33} \\ 
RU & BN & {'BN': 70.0, 'EN': 30.0} \\ 
ES & BN & {\colorbox{red!30}{'EN'} 78.33, 'BN': 21.67} \\
BN & BN & {'BN': 98.33, 'EN': 1.67} \\ 
TH & BN & {\colorbox{red!30}{'EN'} 70.0, 'BN': 30.0} \\
SW & BN & {\colorbox{red!30}{'EN'} 65.0, 'BN': 35.0} \\
TE & BN & {'BN': 55.0, 'EN': 45.0} \\ 
EN & TH & {\colorbox{red!30}{'EN'} 43.81, 'TH': 41.67, 'ZH': 10.24, 'others': 4.29} \\
FR & TH & {\colorbox{red!30}{'EN'} 86.19, 'TH': 11.67, 'ZH': 1.67, 'others': 0.48} \\
DE & TH & {\colorbox{red!30}{'EN'} 95.0, 'TH': 5.0} \\
ZH & TH & {\colorbox{red!30}{'ZH'} 86.67, 'others': 6.43, 'TH': 5.0, 'EN': 1.9} \\
JA & TH & {\colorbox{red!30}{'ZH'} 57.14, 'EN': 28.33, 'TH': 10.0, 'others': 4.52} \\
RU & TH & {\colorbox{red!30}{'EN'} 49.05, 'ZH': 35.0, 'TH': 13.33, 'others': 2.62} \\
ES & TH & {\colorbox{red!30}{'EN'} 83.1, 'ZH': 10.24, 'TH': 3.33, 'others': 3.33} \\
BN & TH & {\colorbox{red!30}{'EN'} 75.0, 'TH': 18.33, 'ZH': 5.0, 'others': 1.67} \\
TH & TH & {'TH': 41.67, 'ZH': 35.71, 'EN': 15.95, 'others': 6.67} \\ 
SW & TH & {\colorbox{red!30}{'EN'} 55.0, 'TH': 45.0} \\
TE & TH & {\colorbox{red!30}{'EN'} 48.33, 'TH': 45.0, 'ZH': 5.0, 'others': 1.67} \\
EN & SW & {\colorbox{red!30}{'EN'} 100.0} \\
FR & SW & {\colorbox{red!30}{'EN'} 100.0} \\
DE & SW & {\colorbox{red!30}{'EN'} 98.33, 'others': 1.67} \\
ZH & SW & {\colorbox{red!30}{'EN'} 100.0} \\
JA & SW & {\colorbox{red!30}{'EN'} 99.76, 'ES': 0.24} \\
RU & SW & {\colorbox{red!30}{'EN'} 100.0} \\
ES & SW & {\colorbox{red!30}{'EN'} 98.33, 'SW': 1.67} \\
BN & SW & {\colorbox{red!30}{'EN'} 96.67, 'SW': 3.33} \\
TH & SW & {\colorbox{red!30}{'EN'} 100.0} \\
SW & SW & {\colorbox{red!30}{'EN'} 75.0, 'SW': 25.0} \\
TE & SW & {\colorbox{red!30}{'EN'} 96.67, 'SW': 3.33} \\
EN & TE & {\colorbox{red!30}{'EN'} 68.33, 'TE': 31.67} \\
FR & TE & {\colorbox{red!30}{'EN'} 61.67, 'TE': 38.33} \\
DE & TE & {\colorbox{red!30}{'EN'} 76.67, 'TE': 23.33} \\
ZH & TE & {'TE': 88.33, 'ZH': 10.24, 'others': 1.43} \\ 
JA & TE & {\colorbox{red!30}{'ZH'} 55.0, 'TE': 28.33, 'others': 7.86, 'EN': 6.9, 'JA': 1.9} \\
RU & TE & {'TE': 95.0, 'EN': 3.33, 'RU': 1.67} \\ 
ES & TE & {'TE': 61.67, 'EN': 38.33} \\ 
BN & TE & {\colorbox{red!30}{'EN'} 65.0, 'TE': 30.24, 'BN': 3.81, 'others': 0.95} \\
TH & TE & {\colorbox{red!30}{'EN'} 76.67, 'TE': 23.33} \\
SW & TE & {'TE': 85.0, 'SW': 8.33, 'EN': 6.67} \\ 
TE & TE & {'TE': 96.67, 'EN': 3.33} \\ 
\bottomrule
\end{tabular}
\caption{Cont. The distribution of the predicted language of the thinking traces, averaged over all questions in the same language, when the LRM is prompted to reason in each thinking language.
}
\label{tab:actual_thinking_language_2}
\end{table}

\section{Details of Post-Training Experiments}
\label{app:post_training}
\subsection{Translated LIMO Examples}
For the post-training in Japanese, Telugu, and Thai, we translate the LIMO \citep{ye2025limoreasoning} into the three languages by \textsc{GPT-4o-mini}. The examples of the translated Japanese instances are illustrated in Table \ref{tab:multi_limo}. For full translated instances in Japanese, Telugu, and Thai, please refer to
\url{https://huggingface.co/collections/shanchen/xreasoning-models-68377e15a2e86143dc4b0383}.

\begin{CJK*}{UTF8}{gbsn}
\begin{table*}[!t]
\centering
\resizebox{0.99\linewidth}{!}{
\begin{tabular}{c|p{3.0cm}|p{22.0cm}|p{1.5cm}}
\toprule
\textbf{Language} & \textbf{Question} & \textbf{Model Response} & \textbf{Answer} \\
\midrule
JA & ルドルフは一定の速度で自転車を漕ぎ、毎マイルの終わりに5分間の休憩を取ります。ジェニファーはルドルフの速度の4分の3の一定の速度で自転車を漕ぎますが、ジェニファーは2マイルの終わりに5分間の休憩を取ります。ジェニファーとルドルフは同時に自転車を漕ぎ始め、$50$マイルの地点にちょうど同じ時間に到着します。彼らは何分かかったのでしょうか？ & <｜Assistant｜><think>

重要な指示：

1. すべてのマークダウン形式をそのまま保持する（*、\#などのマークダウン構文を翻訳しない）。

2. すべての数学的表記を保持する。例えば、\boxed{025}のような表現は変更しない。

3. すべてのコードブロックとその構文を保持する。

4. すべての変数名、数字、技術用語を保持する。

元のテキスト：

さて、この問題をステップバイステップで解決してみましょう。ルドルフとジェニファーが50マイル地点まで自転車を漕いで、同時に到着します。彼らがどれくらいの時間がかかったのかを見つける必要があります。

\textit{[...]}

</think>

Answer: わかりました、これをステップバイステップで分解しましょう。ルドルフとジェニファーがそれぞれの自転車の速度と休憩スケジュールを考慮して、50マイルを自転車で移動するのにかかる総時間を求める必要があります。彼らは一緒に出発し、50マイル地点に同時に到着します。

**ステップ 1: 速度と時間を定義する**

- ルドルフの自転車の速度を \( R \) マイル/分

\textit{[...]}

- **ジェニファーの総時間**:
\[
\frac{200}{3 \times \frac{2}{15}} + 120 = 500 + 120 = 620 \text{ 分}
\]

両者は同時に到着するので、かかった総時間は **620 分** です。

\[\boxed{620}\]
& 620 \\
\midrule
JA & 	
放物線の頂点が $\left(\frac{1}{4},-\frac{9}{8}\right)$ であり、方程式が $y = ax^2 + bx + c$ であるとします。ただし、$a > 0$ であり、$a + b + c$ は整数です。$a$ の最小可能値は $\frac{p}{q}$ の形で表すことができ、ここで $p$ と $q$ は互いに素な正の整数です。 $p + q$ を求めなさい。& <｜Assistant｜><think>

重要な指示：

\textit{[...]}

わかった、だから、頂点が (1/4, -9/8) にあり、方程式が y = ax² + bx + c で、a > 0 かつ a + b + c が整数である放物線の 'a' の最小可能値を見つける必要がある。次に、この最小の 'a' を約分された分数 p/q として表し、p + q を求める。うーん、放物線の頂点形式を思い出すことから始めよう。

\textit{[...]}

</think>

Answer: 最小の可能な値 \(a\) を求めるために、頂点 \(\left(\frac{1}{4}, -\frac{9}{8}\right)\) を持ち、方程式 \(y = ax^2 + bx + c\) で \(a > 0\) かつ \(a + b + c\) が整数である放物線について考えます。まず、放物線の頂点形式から始めます：

\[
y = a\left(x - \frac{1}{4}\right)^2 - \frac{9}{8}
\]

\textit{[...]}

したがって、最小の値 \(a\) は \(\frac{2}{9}\) であり、\(p = 2\) および \(q = 9\) です。したがって、\(p + q = 2 + 9 = 11\)。

\[
\boxed{11}
\]

& 11 \\

\bottomrule
\end{tabular}
}
\caption{The examples of the translated LIMO training instances in Japanese (JA).}
\label{tab:multi_limo}
\end{table*}
\end{CJK*}

\subsection{Post-Trained LRMs and Accessibility}
Overall, seven models are evaluated in our work: the base model, plus two variants (post-trained with 100 or 250 instances) each for Japanese (JA), Telugu (TE), and Thai (TH). A low learning rate (1e-5) is applied to minimize overfitting and catastrophic forgetting during our post-training. All these post-trained LRMs are also publicly accessible via \url{https://huggingface.co/collections/shanchen/xreasoning-681e7625c7a9ec4111a634b6}.

\section{Full Results of Post-Training}
\label{app:full_results}
The full results on the LRMs post-trained with Japanese, Telugu and Thai instances are shown in Table \ref{tab:post-training-ja_full}, \ref{tab:post-training-th_full} and \ref{tab:post-training-te_full}, where the LRMs are always prompt to think in the same language as that of the training data, given queries in different languages. We also include the model performance when the LRMs are post-trained on 500 and 817 Japanese instances.

\begin{table*}[!t]
\centering
\resizebox{0.99\linewidth}{!}{
\begin{tabular}{c|c|c|ccccccccccc}
\toprule
\multirow{2}{*}{\textbf{Eval Metric}} & \multirow{2}{*}{\textbf{\# Ins.}} & \multirow{2}{*}{\textbf{AVG}} & \multicolumn{11}{|c}{\textbf{Query Language}} \\
\cmidrule{4-14}
& & & EN & FR & DE & ZH & JA & RU & ES & BN & TH & SW & TE \\
\midrule
\rowcolor{black!10} \multicolumn{14}{c}{AIME} \\
\midrule
\multirow{4}{*}{\shortstack{\textbf{Matching}\\\textbf{Rate (\%)}}} 
 & None & 41.5 
   & 25.5  & 38.3  & 51.1  & 27.2  & 73.3  & 37.2  & 38.3  & 29.4  & 43.3  & 55.6  & 37.2  \\
 & 100  & 76.4 
   & 95.6  & 93.9  & 72.8  & 94.4  & 93.3  & 92.8  & 93.3  & 11.1  & 65.6  & 68.3  & 58.9  \\
 & 250  & 65.0 
   & 89.4  & 73.9  & 28.3  & 96.7  & 98.9  & 79.4  & 94.9  & 1.7   & 98.3  & 32.2  & 21.7  \\
 & 500 & 74.9 & 92.2 & 73.3 & 84.5 & 94.4 & 93.3 & 77.2 & 86.1 & 31.1 & 93.3 & 48.9 & 49.5 \\
 & 817 & 73.1 & 91.1 & 71.1 & 83.9 & 91.7 & 93.3 & 76.1 & 81.7 & 36.1 & 91.1 & 38.3 & 50.0 \\

\midrule
\multirow{4}{*}{\shortstack{\textbf{Answer}\\\textbf{Acc. (\%)}}} 
 & None & 14.6 
   & 22.2  & 20.5  & 18.9  & 21.1  & 8.3   & 16.1  & 17.2  & 16.1  & 8.8   & 3.9   & 8.3   \\
 & 100  & 4.9 
   & 5.6   & 7.8   & 6.7   & 3.3   & 4.4   & 7.2   & 7.2   & 3.3   & 4.4   & 1.7   & 1.7   \\
 & 250  & 9.0 
   & 6.7   & 8.9   & 20.5  & 5.5   & 4.4   & 8.9   & 6.1   & 21.1  & 5.5   & 0.6   & 10.6  \\
 & 500 & 13.5 & 13.4 & 15.6 & 16.1 & 15.6 & 11.1 & 18.9 & 13.3 & 20.5 & 10.6 & 3.9 & 10.0 \\ 
 & 817 & 12.1 & 13.9 & 12.8 & 12.2 & 13.3 & 13.9 & 13.3 & 14.4 & 15.5 & 11.7 & 2.8 & 8.9 \\

\bottomrule
\rowcolor{black!10} \multicolumn{14}{c}{GPQA} \\
\midrule
\multirow{4}{*}{\shortstack{\textbf{Matching}\\\textbf{Rate (\%)}}} 
 & None & 29.2 
   & 14.5  & 25.3  & 12.3  & 32.3  & 72.4  & 22.9  & 23.4  & 23.7  & 29.8  & 33.3  & 30.5  \\
 & 100  & 61.6 
   & 81.0  & 75.6  & 38.0  & 86.5  & 87.7  & 76.9  & 78.1  & 10.3  & 38.4  & 70.9  & 33.8  \\
 & 250  & 46.8 
   & 67.2  & 57.3  & 2.2   & 80.6  & 92.1  & 48.7  & 60.1  & 0.8   & 77.3  & 20.9  & 7.8   \\
 & 500 & 52.0 & 71.2 & 54.9 & 26.3 & 79.4 & 81.3 & 53.0 & 58.3 & 12.6 & 83.5 & 30.1 & 21.4 \\
 & 817 & 55.7 & 66.5 & 61.1 & 36.6 & 78.3 & 84.0 & 53.7 & 57.6 & 21.7 & 83.7 & 36.2 & 33.6 \\
\midrule
\multirow{4}{*}{\shortstack{\textbf{Answer}\\\textbf{Acc. (\%)}}} 
 & None & 27.4 
   & 34.5  & 29.9  & 30.6  & 28.6  & 23.7  & 28.1  & 29.1  & 28.9  & 24.8  & 21.2  & 22.3  \\
 & 100  & 12.1 
   & 11.4  & 16.5  & 19.4  & 9.6   & 12.8  & 11.1  & 10.6  & 10.4  & 9.5   & 12.6  & 9.8   \\
 & 250  & 9.3 
   & 11.4  & 10.6  & 16.0  & 5.8   & 7.4   & 9.8   & 8.8   & 12.1  & 5.4   & 6.1   & 8.4   \\
 & 500 & 11.0 & 11.9 & 13.1 & 18.5 & 6.9 & 7.3 & 11.6 & 13.3 & 14.3 & 7.8 & 7.1 & 9.6 \\ 
 & 817 & 10.8 & 12.4 & 11.8 & 14.8 & 8.4 & 6.9 & 14.1 & 13.8 & 10.4 & 8.6 & 6.9 & 10.1 \\
\bottomrule
\rowcolor{black!10} \multicolumn{14}{c}{MGSM} \\
\midrule
\multirow{4}{*}{\shortstack{\textbf{Matching}\\\textbf{Rate (\%)}}} 
 & None & 58.6 
   & 80.5  & 58.5  & 25.7  & 67.2  & 93.5  & 68.5  & 65.7  & 39.9  & 36.5  & 44.0  & 64.4  \\
 & 100  & 64.8 
   & 95.6  & 75.9  & 24.5  & 94.9  & 99.2  & 81.7  & 88.1  & 6.7   & 41.2  & 64.4  & 40.9  \\
 & 250  & 77.9 
   & 99.1  & 83.1  & 44.8  & 98.1  & 99.7  & 94.4  & 91.3  & 33.9  & 87.1  & 54.4  & 71.2  \\
 & 500 & 67.5 & 96.4 & 71.7 & 13.6 & 94.5 & 98.1 & 83.5 & 78.5 & 37.3 & 94.5 & 32.3 & 41.6 \\
 & 817 & 70.4 & 97.5 & 73.1 & 28.4 & 93.6 & 96.1 & 84.4 & 80.1 & 40.9 & 89.5 & 36.8 & 53.9 \\
\midrule
\multirow{4}{*}{\shortstack{\textbf{Answer}\\\textbf{Acc. (\%)}}} 
 & None & 50.7 
   & 71.6  & 60.4  & 65.5  & 67.2  & 50.0  & 59.3  & 61.2  & 44.2  & 49.3  & 5.5   & 22.5  \\
 & 100  & 46.4 
   & 66.3  & 54.8  & 63.6  & 59.3  & 51.6  & 58.5  & 57.1  & 40.4  & 46.2  & 3.6   & 11.2  \\
 & 250  & 42.9 
   & 64.3  & 55.3  & 60.4  & 53.5  & 46.1  & 51.9  & 62.5  & 32.8  & 53.7  & 3.6   & 15.3  \\
 & 500 & 44.3 & 67.6 & 55.1 & 65.7 & 48.5 & 38.0 & 54.7 & 59.7 & 42.1 & 31.5 & 2.9 & 20.9 \\
 & 817 & 44.7 & 65.7 & 55.5 & 60.9 & 53.5 & 43.5 & 55.2 & 56.7 & 42.4 & 34.7 & 3.3 & 20.0 \\

\bottomrule
\end{tabular}
}
\caption{Model performance on AIME/GPQA/MGSM dataset with Deepseek-Distill-R1-7B post-trained on 100/250 few Japanese instances. The LRM is prompted to think in Japanese.} 
\label{tab:post-training-ja_full}
\end{table*}

\begin{table*}[!t]
\centering
\resizebox{0.99\linewidth}{!}{
\begin{tabular}{c|c|c|ccccccccccc}
\toprule
\multirow{2}{*}{\textbf{Eval Metric}} & \multirow{2}{*}{\textbf{\# Ins.}} & \multirow{2}{*}{\textbf{AVG}} & \multicolumn{11}{|c}{\textbf{Query Language}} \\
\cmidrule{4-14}
 &  &  & EN & FR & DE & ZH & JA & RU & ES & BN & TH & SW & TE \\
\midrule
\rowcolor{black!10} \multicolumn{14}{c}{AIME} \\
\midrule
\multirow{3}{*}{\shortstack{\textbf{Matching}\\\textbf{Rate (\%)}}}
 & None & 29.1 
   & 17.2  & 24.4  & 30.6  & 9.5   & 25.6  & 25.0  & 25.6  & 33.3  & 39.5  & 54.4  & 35.5  \\
 & 100  & 99.6 
   & 100.0 & 100.0 & 99.4  & 99.4  & 100.0 & 100.0 & 100.0 & 99.4  & 100.0 & 99.4  & 98.3  \\
 & 250  & 91.9 
   & 95.0  & 89.4  & 90.6  & 100.0 & 99.4  & 93.3  & 88.9  & 85.0  & 100.0 & 81.7  & 87.8  \\
 & 500 & 72.1 & 77.8 & 56.7 & 65.0 & 88.3 & 83.9 & 60.6 & 65.5 & 63.3 & 86.1 & 66.1 & 80.0 \\ 
 & 817 & 60.9 & 73.9 & 41.1 & 50.6 & 75.0 & 65.0 & 42.2 & 52.8 & 61.1 & 82.2 & 53.3 & 72.2 \\
\midrule
\multirow{3}{*}{\shortstack{\textbf{Answer}\\\textbf{Acc. (\%)}}}
 & None & 18.6 
   & 29.4  & 25.6  & 25.0  & 21.7  & 14.5  & 23.9  & 22.2  & 13.9  & 14.5  & 3.3   & 11.1  \\
 & 100  & 4.6 
   & 6.7   & 5.0   & 2.2   & 4.5   & 4.4   & 5.5   & 7.2   & 4.4   & 5.6   & 0.6   & 2.8   \\
 & 250  & 8.0 
   & 12.8  & 8.3   & 11.1  & 8.3   & 8.3   & 8.9   & 9.4   & 7.8   & 6.1   & 0.6   & 6.7   \\
 & 500 & 18.8 & 24.4 & 24.5 & 25.5 & 19.4 & 21.1 & 23.3 & 22.8 & 16.1 & 13.3 & 6.7 & 9.5 \\ 
 & 817 & 18.5 & 19.4 & 26.7 & 26.7 & 16.1 & 20.0 & 24.4 & 22.2 & 20.6 & 13.3 & 5.5 & 8.3 \\

\bottomrule

\rowcolor{black!10} \multicolumn{14}{c}{GPQA} \\
\midrule
\multirow{3}{*}{\shortstack{\textbf{Matching}\\\textbf{Rate (\%)}}}
 & None & 25.6 
   & 18.4  & 19.2  & 19.7  & 8.3   & 20.5  & 19.7  & 17.4  & 30.5  & 49.8  & 42.2  & 35.7  \\
 & 100  & 98.0 
   & 99.3  & 98.8  & 97.5  & 99.3  & 99.1  & 99.7  & 99.0  & 96.5  & 99.3  & 98.0  & 91.4  \\
 & 250  & 82.3 
   & 80.5  & 83.8  & 70.5  & 98.5  & 97.5  & 69.7  & 80.8  & 85.2  & 98.3  & 58.6  & 82.2  \\
 & 500 & 47.8 & 53.0 & 44.8 & 37.7 & 86.2 & 59.8 & 24.4 & 51.2 & 39.7 & 79.0 & 14.1 & 36.2 \\ 
 & 817 & 41.8 & 41.7 & 39.9 & 33.3 & 69.7 & 47.2 & 18.9 & 47.3 & 34.5 & 79.3 & 14.7 & 33.3 \\
\midrule
\multirow{3}{*}{\shortstack{\textbf{Answer}\\\textbf{Acc. (\%)}}}
 & None & 27.5 
   & 33.5  & 26.6  & 29.1  & 29.8  & 26.3  & 28.3  & 31.3  & 24.6  & 27.6  & 25.3  & 22.6  \\
 & 100  & 9.1 
   & 10.1  & 8.8   & 9.9   & 6.1   & 10.4  & 9.9   & 9.1   & 10.4  & 11.1  & 7.3   & 7.6   \\
 & 250  & 11.6 
   & 13.3  & 10.9  & 15.5  & 7.1   & 11.3  & 15.3  & 13.0  & 9.4   & 8.6   & 11.6  & 10.6  \\
 & 500 & 18.5 & 20.4 & 20.2 & 23.7 & 14.1 & 16.2 & 24.4 & 21.2 & 17.7 & 14.6 & 16.3 & 14.6 \\ 
 & 817 & 15.4 & 18.7 & 18.0 & 18.4 & 10.1 & 14.6 & 21.2 & 15.3 & 16.5 & 10.4 & 13.6 & 12.5 \\

\bottomrule

\rowcolor{black!10} \multicolumn{14}{c}{MGSM} \\
\midrule
\multirow{3}{*}{\shortstack{\textbf{Matching}\\\textbf{Rate (\%)}}}
 & None & 71.5 
   & 69.9  & 63.5  & 67.6  & 58.4  & 73.3  & 75.0  & 65.9  & 80.7  & 88.5  & 75.5  & 67.7  \\
 & 100  & 97.6 
   & 94.0  & 95.5  & 98.7  & 98.8  & 99.9  & 99.7  & 97.5  & 99.5  & 99.3  & 98.3  & 94.4  \\
 & 250  & 98.3 
   & 99.6  & 94.9  & 94.5  & 99.9  & 99.7  & 99.9  & 97.3  & 99.2  & 99.9  & 98.5  & 98.3  \\
 & 500 & 92.8 & 98.7 & 87.9 & 76.3 & 100.0 & 99.2 & 98.0 & 80.4 & 98.1 & 99.7 & 84.1 & 97.9 \\
 & 817 & 86.0 & 96.1 & 74.9 & 54.0 & 94.4 & 96.1 & 92.1 & 73.2 & 92.7 & 99.1 & 80.3 & 92.5 \\
\midrule
\multirow{3}{*}{\shortstack{\textbf{Answer}\\\textbf{Acc. (\%)}}} 
 & None & 47.9 
   & 69.9  & 55.3  & 57.9  & 65.2  & 38.9  & 57.3  & 62.7  & 35.3  & 50.0  & 5.5   & 21.7  \\
 & 100  & 36.7 
   & 57.7  & 45.6  & 46.8  & 45.9  & 29.5  & 40.7  & 50.5  & 25.0  & 47.2  & 3.1   & 13.5  \\
 & 250  & 45.5 
   & 72.8  & 54.8  & 58.8  & 57.7  & 43.2  & 51.9  & 63.6  & 31.7  & 54.0  & 4.8   & 12.9  \\
 & 500 & 48.1 & 75.1 & 58.0 & 59.6 & 59.9 & 42.8 & 59.9 & 64.3 & 34.1 & 52.9 & 4.1 & 18.1 \\
 & 817 & 45.7 & 71.1 & 56.8 & 58.3 & 58.4 & 42.1 & 55.6 & 61.7 & 30.4 & 49.6 & 2.8 & 15.9 \\

\bottomrule
\end{tabular}
}
\caption{Model performance on AIME/GPQA/MGSM dataset with Deepseek-Distill-R1-7B post-trained on 100/250 few Thai instances. The LRM is prompted to think in Thai.} 
\label{tab:post-training-th_full}
\end{table*}

\begin{table*}[!t]
\centering
\resizebox{0.99\linewidth}{!}{
\begin{tabular}{c|c|c|ccccccccccc}
\toprule
\multirow{2}{*}{\textbf{Eval Metric}} & \multirow{2}{*}{\textbf{\# Ins.}} & \multirow{2}{*}{\textbf{AVG}} & \multicolumn{11}{|c}{\textbf{Query Language}} \\
\cmidrule{4-14}
& & & EN & FR & DE & ZH & JA & RU & ES & BN & TH & SW & TE \\
\midrule
\rowcolor{black!10} \multicolumn{14}{c}{AIME} \\
\midrule
\multirow{3}{*}{\shortstack{\textbf{Matching}\\\textbf{Rate (\%)}}} 
 & None & 15.8  
   & 3.3   & 7.8   & 6.7   & 0.6   & 21.1  & 11.1  & 6.7   & 6.7   & 23.3  & 32.8  & 54.4  \\
 & 100  & 99.6  
   & 99.4  & 98.3  & 100.0 & 99.4  & 100.0 & 100.0 & 100.0 & 99.4  & 98.9  & 100.0 & 100.0 \\
 & 250  & 100.0 
   & 100.0 & 100.0 & 100.0 & 100.0 & 100.0 & 100.0 & 100.0 & 100.0 & 100.0 & 100.0 & 100.0 \\
 & 500 & 99.6 & 100.0 & 98.9 & 100.0 & 100.0 & 99.4 & 100.0 & 100.0 & 100.0 & 98.9 & 99.4 & 99.4 \\ 
 & 817 & 99.9 & 100.0 & 100.0 & 100.0 & 100.0 & 100.0 & 99.4 & 100.0 & 100.0 & 100.0 & 100.0 & 100.0 \\

\midrule
\multirow{3}{*}{\shortstack{\textbf{Answer}\\\textbf{Acc. (\%)}}} 
 & None & 24.3  
   & 31.7  & 33.3  & 28.3  & 27.8  & 25.6  & 28.3  & 30.6  & 23.3  & 23.9  & 5.6   & 8.9   \\
 & 100  & 0.2   
   & 1.1   & 0.0   & 0.0   & 0.0   & 0.6   & 0.0   & 0.0   & 0.0   & 0.0   & 0.6   & 0.0   \\
 & 250  & 0.8   
   & 0.6   & 0.0   & 1.1   & 1.7   & 1.1   & 1.1   & 0.6   & 0.6   & 0.6   & 0.0   & 0.6   \\
 & 500 & 3.3 & 3.3 & 4.4 & 5.0 & 3.9 & 2.2 & 2.8 & 4.4 & 3.3 & 2.8 & 0.0 & 4.4 \\ 
 & 817 & 2.9 & 4.4 & 2.2 & 5.0 & 5.6 & 4.4 & 2.2 & 1.1 & 0.6 & 2.2 & 0.6 & 3.3 \\

\bottomrule
\rowcolor{black!10} \multicolumn{14}{c}{GPQA} \\
\midrule
\multirow{3}{*}{\shortstack{\textbf{Matching}\\\textbf{Rate (\%)}}} 
 & None & 5.4   
   & 0.0   & 2.0   & 0.0   & 0.3   & 18.4  & 1.5   & 0.5   & 0.7   & 8.1   & 1.3   & 26.8  \\
 & 100  & 99.2  
   & 98.5  & 98.0  & 98.5  & 99.5  & 99.7  & 99.7  & 97.7  & 99.2  & 99.8  & 99.3  & 100.0 \\
 & 250  & 99.8  
   & 99.8  & 99.8  & 100.0 & 99.8  & 99.7  & 100.0 & 99.8  & 99.7  & 99.8  & 99.7  & 100.0 \\
\midrule
\multirow{3}{*}{\shortstack{\textbf{Answer}\\\textbf{Acc. (\%)}}} 
 & None & 33.5  
   & 41.9  & 37.7  & 38.7  & 33.3  & 26.8  & 34.4  & 36.0  & 32.6  & 32.6  & 25.2  & 24.6  \\
 & 100  & 3.0   
   & 2.8   & 3.7   & 2.7   & 2.0   & 2.2   & 3.0   & 3.2   & 2.5   & 3.0   & 4.0   & 4.4   \\
 & 250  & 1.6   
   & 1.7   & 2.0   & 1.5   & 1.0   & 1.7   & 1.7   & 1.8   & 0.7   & 1.3   & 1.8   & 1.6   \\
 & 500 & 1.5 & 2.3 & 1.5 & 1.7 & 1.2 & 1.5 & 1.0 & 1.2 & 1.7 & 1.7 & 1.8 & 1.3 \\ 
 & 817 & 0.6 & 0.7 & 1.2 & 0.7 & 0.5 & 0.3 & 0.5 & 1.0 & 0.7 & 0.7 & 0.5 & 0.3 \\

\bottomrule
\rowcolor{black!10} \multicolumn{14}{c}{MGSM} \\
\midrule
\multirow{3}{*}{\shortstack{\textbf{Matching}\\\textbf{Rate (\%)}}} 
 & None & 94.6  
   & 97.3  & 98.3  & 94.8  & 87.5  & 99.1  & 97.7  & 93.7  & 77.7  & 97.3  & 96.7  & 100.0 \\
 & 100  & 99.6  
   & 99.5  & 99.5  & 99.5  & 99.9  & 99.9  & 100.0 & 99.9  & 99.3  & 99.9  & 99.2  & 100.0 \\
 & 250  & 99.9  
   & 100.0 & 99.6  & 99.6  & 100.0 & 99.7  & 100.0 & 99.9  & 99.2  & 100.0 & 99.7  & 100.0 \\
 & 500 & 99.9 & 99.9 & 99.9 & 99.5 & 100.0 & 100.0 & 99.9 & 99.9 & 99.6 & 100.0 & 99.9 & 100.0 \\
 & 817 & 99.7 & 100.0 & 99.9 & 99.6 & 100.0 & 99.6 & 100.0 & 99.7 & 99.3 & 99.7 & 99.3 & 99.9 \\

\midrule
\multirow{3}{*}{\shortstack{\textbf{Answer}\\\textbf{Acc. (\%)}}} 
 & None & 31.7  
   & 58.3  & 37.1  & 38.3  & 35.9  & 20.1  & 36.8  & 39.2  & 35.5  & 20.4  & 3.4   & 24.4  \\
 & 100  & 5.5   
   & 9.3   & 5.6   & 5.6   & 5.3   & 3.3   & 3.7   & 7.5   & 6.3   & 3.1   & 0.4   & 10.8  \\
 & 250  & 7.8   
   & 12.9  & 11.2  & 9.3   & 8.1   & 5.3   & 7.6   & 9.6   & 8.4   & 4.8   & 1.1   & 10.1  \\
 & 500 & 17.3 & 31.6 & 24.7 & 21.9 & 18.4 & 13.2 & 17.6 & 22.8 & 13.9 & 11.1 & 0.3 & 15.2 \\
 & 817 & 13.1 & 21.3 & 15.7 & 18.9 & 14.4 & 10.7 & 15.7 & 18.1 & 11.5 & 6.7 & 0.3 & 10.9 \\

\bottomrule
\end{tabular}
}
\caption{Model performance on AIME/GPQA/MGSM dataset with Deepseek-Distill-R1-7B post-trained on 100/250 few Telugu instances. The LRM is prompted to think in Telugu.} 
\label{tab:post-training-te_full}
\end{table*}


\begin{table*}[!t]
\centering
\resizebox{0.99\linewidth}{!}{
\begin{tabular}{c|c|c|ccccccccccc}
\toprule
\multirow{2}{*}{\textbf{Eval Metric}} & \multirow{2}{*}{\textbf{\# Ins.}} & \multirow{2}{*}{\textbf{AVG}} & \multicolumn{11}{|c}{\textbf{Query Language}} \\
\cmidrule{4-14}
& & & EN & FR & DE & ZH & JA & RU & ES & BN & TH & SW & TE \\
\midrule
\rowcolor{black!10} \multicolumn{14}{c}{AIME} \\
\midrule
\multirow{3}{*}{\shortstack{\textbf{Matching}\\\textbf{Rate (\%)}}} 
 & None & 41.5 
   & 25.5  & 38.3  & 51.1  & 27.2  & 73.3  & 37.2  & 38.3  & 29.4  & 43.3  & 55.6  & 37.2  \\
 & 250  & 65.0 
   & 89.4  & 73.9  & 28.3  & 96.7  & 98.9  & 79.4  & 94.9  & 1.7   & 98.3  & 32.2  & 21.7  \\
 & MG-250 & 68.7
 & 64.4 & 65.5 & 75.6 & 67.8 & 91.1 & 69.5 & 71.1 & 48.9 & 69.4 & 67.8 & 65.0\\
\midrule
\multirow{3}{*}{\shortstack{\textbf{Answer}\\\textbf{Acc. (\%)}}} 
 & None & 14.6 
   & 22.2  & 20.5  & 18.9  & 21.1  & 8.3   & 16.1  & 17.2  & 16.1  & 8.8   & 3.9   & 8.3   \\
 & 250  & 9.0 
   & 6.7   & 8.9   & 20.5  & 5.5   & 4.4   & 8.9   & 6.1   & 21.1  & 5.5   & 0.6   & 10.6  \\
 & MG-250 & 7.9
 & 11.1 & 11.6 & 12.2 & 8.3 & 2.8 & 11.1 & 10.5 & 10.0 & 3.9 & 1.7 & 3.3\\
\bottomrule
\rowcolor{black!10} \multicolumn{14}{c}{GPQA} \\
\midrule
\multirow{3}{*}{\shortstack{\textbf{Matching}\\\textbf{Rate (\%)}}} 
 & None & 29.2 
   & 14.5  & 25.3  & 12.3  & 32.3  & 72.4  & 22.9  & 23.4  & 23.7  & 29.8  & 33.3  & 30.5  \\
 & 250  & 46.8 
   & 67.2  & 57.3  & 2.2   & 80.6  & 92.1  & 48.7  & 60.1  & 0.8   & 77.3  & 20.9  & 7.8   \\
 & MG-250 & 55.7
 & 53.2 & 61.1 & 30.3 & 79.0 & 86.2 & 51.1 & 59.1 & 35.2 & 47.3 & 60.9 & 49.0 \\
\midrule
\multirow{3}{*}{\shortstack{\textbf{Answer}\\\textbf{Acc. (\%)}}} 
 & None & 27.4 
   & 34.5  & 29.9  & 30.6  & 28.6  & 23.7  & 28.1  & 29.1  & 28.9  & 24.8  & 21.2  & 22.3  \\
 & 250  & 9.3 
   & 11.4  & 10.6  & 16.0  & 5.8   & 7.4   & 9.8   & 8.8   & 12.1  & 5.4   & 6.1   & 8.4   \\
 & MG-250 & 21.6
 & 24.2 & 24.9 & 23.9 & 21.4 & 20.5 & 23.4 & 24.3 & 21.4 & 20.5 & 18.0 & 15.1\\
\bottomrule
\rowcolor{black!10} \multicolumn{14}{c}{MGSM} \\
\midrule
\multirow{3}{*}{\shortstack{\textbf{Matching}\\\textbf{Rate (\%)}}} 
 & None & 58.6 
   & 80.5  & 58.5  & 25.7  & 67.2  & 93.5  & 68.5  & 65.7  & 39.9  & 36.5  & 44.0  & 64.4  \\
 & 250  & 77.9 
   & 99.1  & 83.1  & 44.8  & 98.1  & 99.7  & 94.4  & 91.3  & 33.9  & 87.1  & 54.4  & 71.2  \\
 & MG-250 & 67.3
 & 91.6 & 68.8 & 30.1 & 81.5 & 97.5 & 77.3 & 79.5 & 41.5 & 45.9 & 57.3 & 68.9 \\
\midrule
\multirow{3}{*}{\shortstack{\textbf{Answer}\\\textbf{Acc. (\%)}}} 
 & None & 50.7 
   & 71.6  & 60.4  & 65.5  & 67.2  & 50.0  & 59.3  & 61.2  & 44.2  & 49.3  & 5.5   & 22.5  \\
 & 250  & 42.9 
   & 64.3  & 55.3  & 60.4  & 53.5  & 46.1  & 51.9  & 62.5  & 32.8  & 53.7  & 3.6   & 15.3  \\
 & MG-250 & 50.5
 & 72.5 & 58.8 & 61.5 & 65.6 & 56.0 & 59.5 & 63.3 & 44.3 & 48.4 & 4.7 & 21.1 \\
\bottomrule
\end{tabular}
}
\caption{Model performance on AIME/GPQA/MGSM dataset with Deepseek-Distill-R1-7B post-trained on 250 few Japanese instances and the merged LRM MG-250 whose parameters are the weighted average of the original model with 0.25 times the parameter of the LRM post-trained with 250 Japanese instances. All LRMs are prompted to think in Japanese.} 
\label{tab:post-training-ja_full_merge}
\end{table*}

\end{document}